\documentclass[sigconf]{acmart}

\AtBeginDocument{%
  \providecommand\BibTeX{{%
    \normalfont B\kern-0.5em{\scshape i\kern-0.25em b}\kern-0.8em\TeX}}}

\usepackage{subcaption}
\usepackage{longtable}
\usepackage{colortbl}
 \setcopyright{none} 

\copyrightyear{2021}
\acmYear{2021}
\setcopyright{acmlicensed}
\acmConference[GEOAI '21]{4th ACM SIGSPATIAL International Workshop on AI for Geographic Knowledge Discovery }{November 2--5, 2021}{Beijing, China}
\acmBooktitle{4th ACM SIGSPATIAL International Workshop on AI for Geographic Knowledge Discovery (GEOAI '21), November 2--5, 2021, Beijing, China}
\acmPrice{15.00}
\acmDOI{10.1145/3486635.3491076}
\acmISBN{978-1-4503-9120-7/21/11}

\begin{document}

\title{hex2vec - Context-Aware Embedding H3 Hexagons with OpenStreetMap Tags}

\author{Szymon Woźniak}
\email{swozniak6@gmail.com}
\orcid{0000-0002-2047-1649}
\affiliation{%
  \institution{Department of Artificial Intelligence,\\ Wrocław University of Science and Technology}
  \city{Wrocław}
  \country{Poland}}

\author{Piotr Szymański}
\email{piotr.szymanski@pwr.edu.pl}
\orcid{0000-0002-7733-3239}
\affiliation{%
  \institution{Department of Artificial Intelligence,\\ Wrocław University of Science and Technology}
  \city{Wrocław}
  \country{Poland}}
\renewcommand{\shortauthors}{Woźniak and Szymański}

\begin{abstract}
Representation learning of spatial and geographic data is a rapidly developing field which allows for similarity detection between areas and high-quality inference using deep neural networks. Past approaches however concentrated on embedding raster imagery (maps, street or satellite photos), mobility data or road networks. In this paper we propose the first approach to learning vector representations of OpenStreetMap regions with respect to urban functions and land-use in a micro-region grid. We identify a subset of OSM tags related to major characteristics of land-use, building and urban region functions, types of water, green or other natural areas. Through manual verification of tagging quality, we selected 36 cities were for training region representations. Uber's H3 index was used to divide the cities into hexagons, and OSM tags were aggregated for each hexagon. We propose the hex2vec method based on the Skip-gram model with negative sampling. The resulting vector representations showcase semantic structures of the map characteristics, similar to ones found in vector-based language models. We also present insights from region similarity detection in six Polish cities and propose a region typology obtained through agglomerative clustering.
\end{abstract}




\begin{teaserfigure}
  \includegraphics[width=\textwidth]{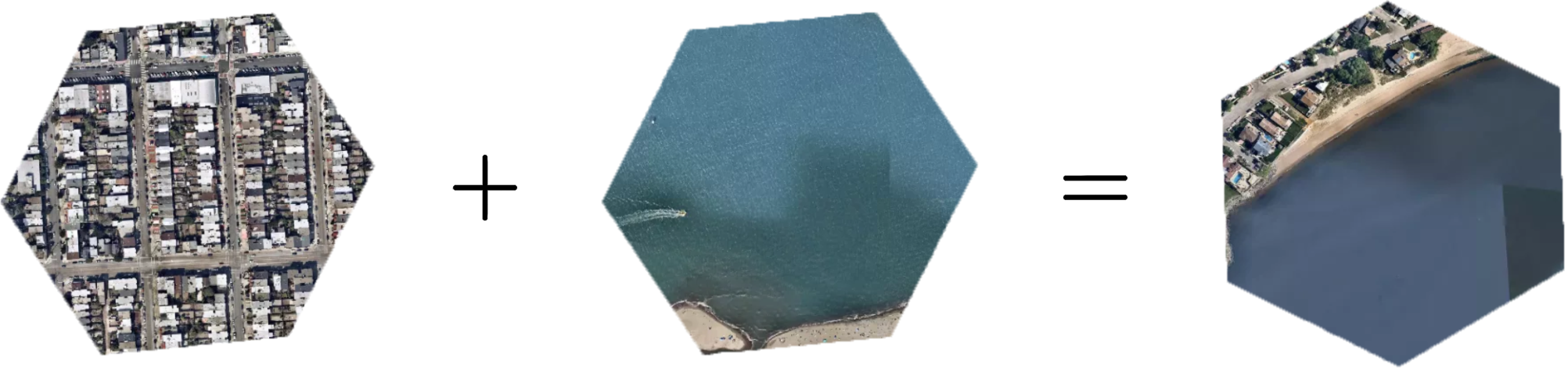}
  \caption{Vector addition in the hex2vec embedding space.}
  \Description{Adding a region with a coast to a densely built-up region results in a region with buildings along the coast.}
  \label{fig:teaser}
\end{teaserfigure}

\maketitle

\section{Introduction}
Nowadays, spatial data is generated and gathered in large quantities in many cities and urban areas around the world. The increasing availability of data makes its use in automating analyses, planning and decision-making possible on an ever-larger scale.
This type of data can be used for many urban tasks, such as estimating travel times, recommending next points of interest (POI), estimating traffic volumes or detecting region functions. Automating some of these processes could significantly simplify the work of city planners and officials. It could also allow decisions to be made based on data rather than beliefs and intuitions. The ability to compare regions as well as entire cities could also allow planning decisions to be directed towards achieving a particular style or desired characteristics.

However, the application of automation methods is often limited by two factors.
The first problem comes from the availability of urban spatial data. Some of it is gathered by authorities and private companies and is not made available to the general public.
The second problem arises from the use of solutions custom-tailored to the specific data format or specific researchers' need. While solutions like that may outperform other more general ones, they can be applied to only one specific use case.

In this paper, effort will be made to propose a more general approach to using urban spatial data in machine learning solutions. It will use publicly available data from OpenStreetMap enabling the application of this solution in any region for which data has been entered in this system. It will be based on embedding city regions in vector space allowing straightforward comparison of regions and easy further application of existing machine learning methods.

In this paper we follow the path of the original, groundbreaking word2vec paper. Our main contribution is that we are the first to show that skip-gram approaches allow obtaining vector representations of OpenStreetMap data which exhibit strong semantically interpretable characteristics. We show that by providing insight into:
\begin{itemize}
    \item how regions cluster based on vector similarity,
    \item how vector arithmetic approaches follow semantic interpretation,
    \item how semantic meaning is exhibited when interpolating between two vectors.
\end{itemize}

The aim of this paper, is not to claim state-of-the-art results on downstream tasks, but rather to provide the community with new representation methods which can aid solving downstream tasks. We provide the code and obtained vectors in an accompanying repository\footnote{\url{https://github.com/pwr-inf/hex2vec}}.

\section{Background}

\subsection{OpenStreetMap functional information}
One possible source of spatial data is functional information of objects, available in OpenStreetMap (OSM). It is a free, editable map of the whole world, built by volunteers and released with an open-content license\cite{osm-about}. The physical world is modelled in OSM using three types of elements:
\begin{itemize}
    \item nodes - defining points in space,
    \item ways - defining linear features and boundaries,
    \item relations - structures used to group other features into bigger objects.
\end{itemize}
Each of these elements can have one or more tags associated with it. They come in the form of \textit{key=value} pairs and are used to describe the meaning of the element. The \textit{key=value} pairs are what is referred to as \textit{functional information} in the thesis title. They carry information about buildings, land use, amenities and much more. When combined, they should hopefully represent city-regions well.

%\begin{figure}[htpb]
%    \centering
%    \includegraphics[width=.48\textwidth]{images/osm_example.png}
%    \caption{Example relation from OpenStreetMap, representing a university building. }
%    \label{fig:osm-example}
%\end{figure}

%\begin{figure}[htpb]
%    \centering
%    \includegraphics[width=.48\textwidth]{images/osm_rynek_wroclaw.png}
%    \caption{Elements from a region in Wrocław, with \textit{amenity} key attached.}
%    \label{fig:osm-rynek-wroclaw}
%\end{figure}

When working with OpenStreetMap data, one can face at least two problems. The first one is the mapping coverage. The map elements are abundant in certain areas, however, that is not the case for every region in the world. As stated in \cite{osm-about} \textbf{the map is not finished yet}. That is why even when working with highly urbanized areas, such as many European capital cities, one should be careful not to include a poorly mapped region.\\
The second possible problem can be caused by the fact that there is no fixed set of tags in OpenStreetMap. Instead, it is based on conventions, but those are not enforced in any strict way. The tags should therefore be carefully filtered to include only relevant information.\\
These problems will be discussed in more detail in the section related to data.

\subsection{Microregions and spatial indexes}\label{ch:spatial-indexes}
\begin{figure}[htpb]
    \centering
    \includegraphics[width=0.48\textwidth]{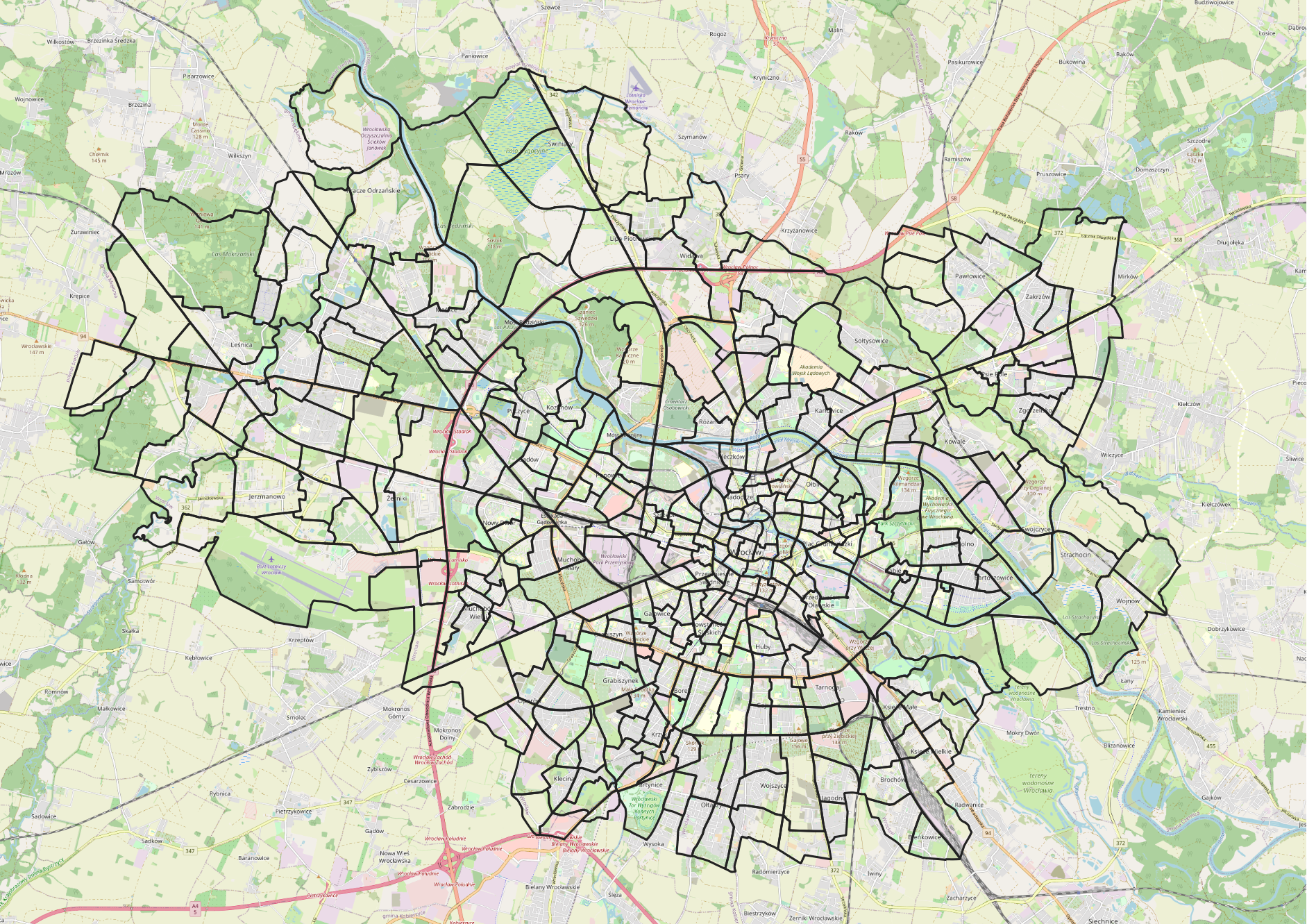}
    \caption{Example of manually drawn regions. Used in comprehensive traffic study in Wrocław, Poland in 2018. Figure created using data from \cite{kbr}.}
    \label{fig:kbr-regions}
\end{figure}

One important question that comes to mind when reading the topic of the thesis is how space should be divided into \textit{microregions}. Dividing the space into regions manually may be a good choice if one wants to include the domain knowledge they possess into the prepared solution. This approach is however almost impossible to apply when the author does not have excellent knowledge of the city under analysis. It becomes all the more problematic when the number of studied cities increases. Then it is even harder for one to have all the required knowledge, and the manual work becomes tedious if not infeasible. An example of manual partitioning is presented in Figure \ref{fig:kbr-regions}. The breakdown was prepared as part of a comprehensive traffic study conducted in Wrocław, Poland in 2018. Another approach that could be taken is to use regional divisions that already exist, such as postal code regions or Zip Code Tabulation Areas, used by United States Census Bureau. The main problem with using postal code areas however is the fact that they are of very different shapes and sizes, which makes analysis more difficult. They may also change for reasons unrelated to the conducted study.

\paragraph{Using spatial indexes}
The solutions presented earlier have drawbacks that hinder their use in automated analyses of large data sets. An approach that can be used to avoid them is to use spatial indexes based on discrete global grids. These are structures that divide the earth's surface into discrete parts. They allow regions to be uniquely indexed by individual numbers. As such, they can be used as database indexes to significantly speed up queries. Some of them are also hierarchical, meaning that they decompose the surface into multiple levels of cells. Two such indexes are Uber’s Hexagonal Hierarchical Spatial Index (H3)\cite{uber-h3}, and Google's S2\cite{google-s2}. They share some similar features, but as H3 has advantages relevant to this work, it will be discussed here in more detail. H3 spatial index uses hexagons as the base shape for dividing the space. Two other regular polygons can tessellate (cover a plane using with no overlaps and no gaps). Hexagons, however, have an important advantage: they share an edge with all their neighbors, and the distance between the centre of the hexagon and the centres of each of its neighbors is the same. It greatly simplifies the analyses, as there is no need to distinguish between edge and vertex neighbors. 

\subsection{Embedding}
\begin{figure}
    \centering
    \includegraphics[width=0.48\textwidth]{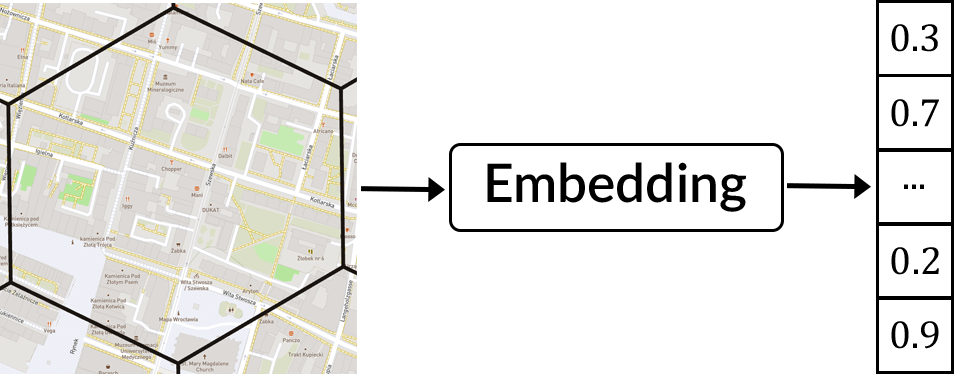}
    \caption{Simplified view of the embedding method's goal.}
    \label{fig:embedding}
\end{figure}

Another important term used in the title of the thesis is \textit{embedding}. It is a method from representation learning, that aims to automatically learn a mapping function $f: O \mapsto \mathbb{R}^d$ from an object of interest to its vector representation. The obtained representation can then be used in many different tasks, with the use of methods that operate on vectors. Recently, this kind of approach has proven to be very effective in the area of natural language processing. In tasks from this field, vector representations are created for words \cite{word2vec}\cite{glove}, sentences \cite{s-bert} or documents \cite{doc2vec} with the usage of language models, trained on large corpora of texts. However, it is yet unclear how to effectively apply similar techniques to spatial data. The goal would be to generate a representation for each analyzed region. A simplified diagram of the embedding process is shown in Figure \ref{fig:embedding}.

Loc2Vec\cite{loc2Vec} is one of the first works that tackled geospatial data embedding. In Loc2Vec, the problem of embedding spatial data was approached by drawing inspiration from the computer vision domain. The authors needed to generate representations for rectangular regions around a point of given coordinates. Their solution was based on rasterizing data from OpenStreetMap and then working on that image. They stated that they used images because at the moment it was unclear how to use the large amount of unstructured data from OSM. In an attempt to keep more information about each object, they did not rasterize the data to an RGB image, as it would flatten the data and make many symbols overlap. Instead, they created a 12-channel tensor, where each channel represented a separate type of object from OSM such as road networks, land cover or amenities. As the authors worked with images, they chose to use a custom convolutional neural network as their embedding model. To avoid manually labelling the data, they decided to train it using a self-supervised approach with triplet loss \cite{triplet-networks}. To create the triplets, the authors made use of what is called the \textit{first law of geography}. It states that `Everything is related to everything else, but near things are more related than distant things \cite{first-law}. Following this assumption, the authors created positive examples for triplets, by shifting and rotating the area of interest, but only so that the areas were still overlapping. The negative samples were created by randomly sampling from the rest of the mini-batch. The obtained embedding vectors, allowed to perform meaningful interpolations and vector arithmetic in the embedding space. 

The approach taken in Tile2Vec\cite{tile2vec} is very similar to the one shown in Loc2Vec. Once again, there is a reference to \textit{the first law of geography}. It is presented as analogous to the distributional hypothesis from natural language processing. The authors decided to use a convolutional neural network as an embedding method, they trained it using the triplet loss and used the same sampling procedure.  There are two main differences between this work and Loc2Vec that are worth pointing out. The first one is the type of images in the dataset. Here, instead of rasters from OpenStreetMap, satellite imagery was used. The second difference is that they used the ResNet-18 architecture \cite{resnet}. The authors carried out more extensive testing of model quality, along with the use of learned embeddings in several downstream tasks. 

Another work that aims to find the embedding of regions in a city is Zone2Vec\cite{zone2vec}. In this work, the authors use human mobility data in the form of trajectories generated by Beijing taxis. Regions in the city were automatically derived using road networks. Afterwards, the trajectories from taxis were transformed into sequences of regions. To obtain region representations, the authors assumed that regions in sequences behave similarly to words in sentences. Then, using the Skip-Gram model\cite{word2vec}, they maximized the average log probability of regions given context regions from the sequences. The negative samples were chosen at random from outside the neighborhood of the current region. Additionally social network data was used to create a document-topic semantic matrix via the Doc2Vec\cite{doc2vec} embedding. The method was evaluated on two tasks: a supervised zone classification task using a one-vs-rest classifier trained on embeddings as input, and unsupervised region similarity detection using spectral clustering applied to the embeddings.

One relatively complex embedding method proposed so far is RegionEncoder\cite{regionencoder}. The authors used multiple multimodal data sources to create the region embeddings. The data consisted of mobility from taxis, categories of points-of-interest (POI) and satellite images. The first step of the embedding was splitting the area into regions automatically by partitioning the analyzed space into rectangles of selected width and height. After that, each type of data was preprocessed separately. The images were simply cut out using bounding boxes defined by the regions. The POI categories were counted and then normalized, forming a POI category distribution vector for each region. The taxi mobility data was transformed into a graph. Each trip's origin and destination were substituted by the corresponding regions. Then a mobility flow graph was constructed where the vertices represented the regions and its weighted edges represented the normalized counts of taxi trips between them. The model used for embedding consisted of several parts. The first one was a denoising convolutional autoencoder. The second one was a graph convolutional neural network. It was applied to mobility and POI data. These parts produced two embedding vectors $h_i$ and $h_g$ respectively. The last part of the model was a multi-layer perceptron (MLP). It was used to merge the previously generated embeddings. It acted as a discriminator, aiming to tell if $h_i$ and $h_g$ come from the same region.  The authors evaluated the embeddings from RegionEncoder on region popularity prediction and house price prediction downstream tasks. They were able to outperform a multitude of baseline models, including ones specialized for geospatial data like Tile2Vec\cite{tile2vec}. The authors also evaluated the latent space learned by the model, by showing the nearest neighbors of three selected regions. They show that the nearest regions are similar both visually and in terms of frequent POI categories.

Urban2Vec\cite{urban2vec} is another example of a relatively complex approach, aiming at creating region embeddings using multimodal geospatial data. As was the case with many methods, the authors used POI data, but what is more interesting is that they also used Street View Imagery. To define regions authors utilized the existing division of census tracts. The embedding method consists of several steps. The embedding vectors for Street View images are created using a convolutional network learned with triplet loss. The initial regions' embeddings are created as the average of their respective Street View images. In the last step, textual data related to POI data are taken into account. These come from categories, their ratings and user reviews. The vectors for these words are obtained from the GloVe\cite{glove} model. The triplet loss is then used again to bring the region embedding close to its POI words' embeddings. The method was evaluated on several regression tasks. In the end, the authors also group the embeddings using K-means clustering and explore the learned space by searching for similar neighborhoods in different cities.

Another work that undertakes region embedding is Region2Vec. The publication analyzed the Sanya city area in China. The authors used data on POIs and from a mobile operator. Regions in the city were separated based on base station locations using a Voronoi diagram. To embed the POI data, the authors used GloVe and LDA models. Training data for these methods were created by treating base stations as documents and POIs as words. However, the authors did not propose a method to create a global embedding of regions. Instead, they generated a similarity matrix for each type of embedding using Pearson correlation. They then averaged these matrices obtaining a global similarity measure of regions. To evaluate the embeddings, the authors performed region clustering using the K-means method. 

\citet{wang2019learning} propose IRN2Vec - a neural network based approach to constructing vector representations of road intersections using OSM data. The representation is learned from attributes present in the tags containing the highway primary key in OSM data. Such tags denote can denote among others traffic signals, signs, indicate speed camera presence and the type of junction. This approach is shown to be beneficial for road network related tasks such as classifying whether a given segment includes a signal or a pedestrian crossing; or travel time estimation tasks. It's goal is however different than the goal of this paper - as they do not aim to model the entirety of the urban area.

The general approach is very similar, across most of the mentioned works. Loc2Vec\cite{loc2Vec} and Tile2Vec\cite{tile2vec} seem to have created a paradigm of their own that was often used later. The authors first assume that \textit{the first law of geography} holds either explicitly or implicitly. Then they use this assumption to train a model using triplet loss\cite{loc2Vec, tile2vec, urban2vec}, Skip-gram loss\cite{zone2vec, regionencoder} and other training procedures drawn from more developed areas such as natural language processing\cite{region2vec} or network analysis\cite{node2vec}. The types of data used in the mentioned papers included check-ins, mobility data, OpenStreetMap rasters, satellite images, and Street View images. Unfortunately, only one work using OpenStreetMap data was found\cite{loc2Vec}, and it also treated it as images. As stated before, many works presented here used data that is not publicly available, such as mobility data from taxi companies\cite{zone2vec, node2vec, regionencoder} or mobile operators\cite{region2vec}. These methods are therefore significantly harder to apply on large scale, due to problems with data availability. Most of the presented works used a simple grid-based method to divide the regions. Other approaches included dividing the city based on the city's road network\cite{zone2vec}, using the existing division from census tracts\cite{urban2vec} and a Voronoi diagram created using base stations' locations. No work for learning OSM vector representations using spatial indexes was found, which is an additional element of the novelty of our contribution. 

\begin{table}[htpb]
\centering
\caption[Characteristics of the mentioned works.]{Characteristics of the mentioned works. The data columns from left to right mean: \textbf{CI} - Check-ins, \textbf{M} - Mobility, \textbf{OR/T} - OpenStreetMap  (OR) or tags (OT), \textbf{P} - Points-of-interest, \textbf{SAT} - Satellite imagery, \textbf{SV} - Street view. The \textbf{Sim} column is used to indicate that the work has undertaken the task of similarity detection of regions at the city level or above (not just pairwise). The \textbf{Div} column specifies the method of dividing the regions. The individual values indicate: \textbf{G} - rectangular grid, \textbf{R} - road networks, \textbf{E} - existing division, \textbf{V} - Voronoi diagram, \textbf{H} hexagonal grid.}
\label{tab:paper-features}
\begin{tabular}{@{}lcccccccc@{}}
\toprule
\multicolumn{1}{c}{\textbf{}}       & \multicolumn{6}{c}{\textbf{Data used}}                                                                          & \multicolumn{1}{l}{}             & \multicolumn{1}{l}{}             \\ \cmidrule(lr){2-7}
\multicolumn{1}{c}{\textbf{Method}} & \textbf{CI} & \textbf{M} & \multicolumn{1}{l}{\textbf{OR/T}} & \textbf{P} & \textbf{SAT} & \textbf{SV} & \multicolumn{1}{l}{\textbf{Sim}} & \multicolumn{1}{l}{\textbf{Div}} \\ \midrule
Loc2Vec                             &             &              & OR                                 &              &              &             & \checkmark                                & G                                \\
Tile2Vec                            &             &              &                                   &              & \checkmark            &             &                                  & G                                \\
Zone2Vec                            & \checkmark           & \checkmark            &                                   & \checkmark            &              &             & \checkmark                                & R                                \\
RegionEncoder                       &             & \checkmark            &                                   & \checkmark            & \checkmark            &             &                                  & G                                \\
Urban2Vec                           &             &              &                                   & \checkmark            &              & \checkmark           & \checkmark                                & E                                \\
Region2Vec                          &             & \checkmark            &                                   &              &              &             & \checkmark                                & V                                \\ 
IRN2Vec                          &             &             &                                   OT &              &              &             &                                 & R                                \\ \hline
Hex2Vec                           &             & & OT                                               &              &              &             &                                 & H                                \\ 
\bottomrule
\end{tabular}
\end{table}

\section{Data}\label{ch:data}
We used publicly available data from OpenStreetMap. All of the data used to prepare the final analyses were downloaded on 17-18.05.21 using the OSMnx library\cite{osmnx}. The library, besides other features, provides an easy-to-use wrapper around Nominatim and Overpass APIs of OpenStreetMap. The first one allows for translating natural language queries into addresses or bounding polygons. The second one allows querying the OpenStreetMap databases to get selected parts of data. One could for example query it to get all the geometries with the \textit{building} key attached, to get all buildings from a selected region.  The data in OpenStreetMap consists of elements representing real-world features through geometries. Each element can have multiple tags (key=value pairs) attached to specify its meaning. As the list of tags is not closed, and anyone could input something arbitrary into the system, it was truncated to include only the keys and values listed on the OSM wikipedia page\cite{osm-wiki-features}. The full list is available in Table \ref{tab:osm-keys}.

\begin{figure}[htpb]
    \centering
    \includegraphics[width=0.48\textwidth]{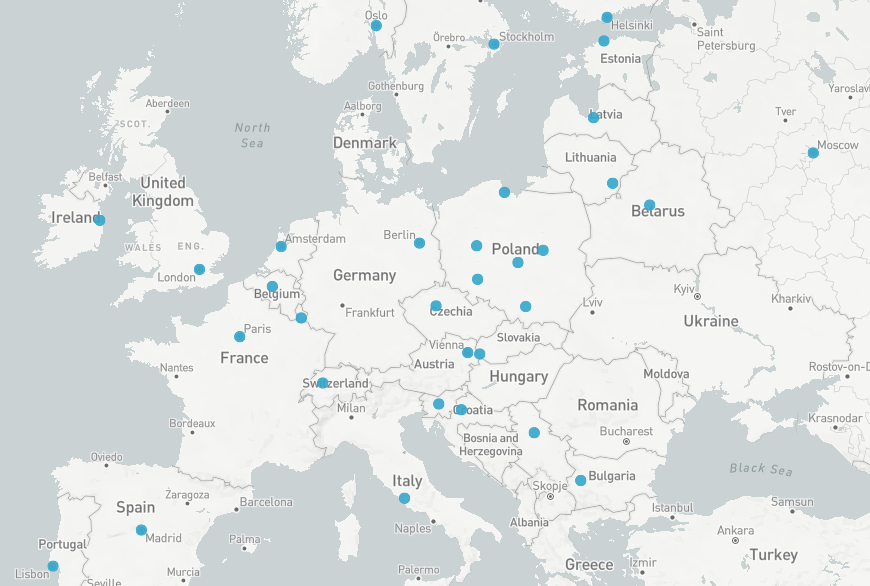}
    \includegraphics[width=0.48\textwidth]{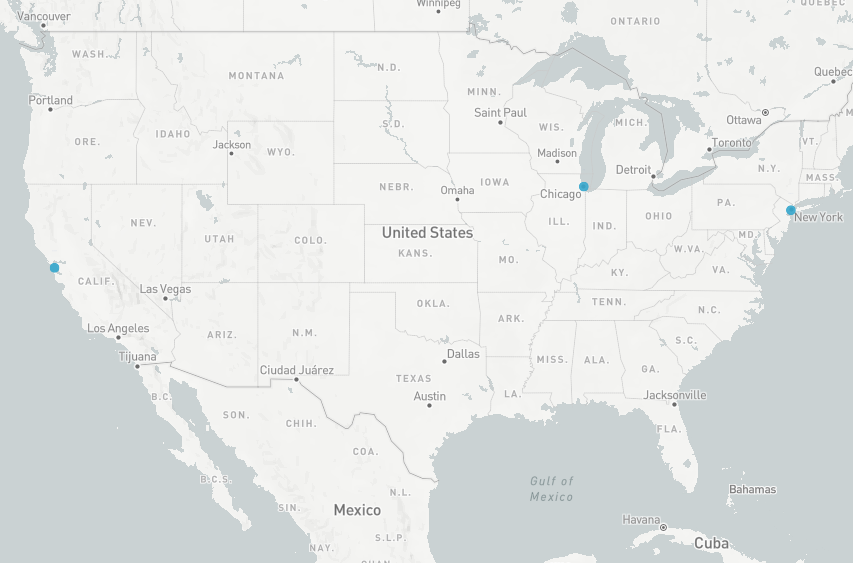}
    \caption[Cities selected for analysis.]{Cities selected for analysis. Nur-Sultan and Reykjavík were omitted here to improve readability.}
    \label{fig:cities}
\end{figure}

\subsection{Selection of cities}
The list of cities considered in the construction of the dataset included ones of different urban structures, sizes and levels of development. Overall, 56 cities including European capitals, 6 additional largest cities in Poland by population, 5 cities from the United States and the capital city of Russia - Moscow, were taken into consideration. For each of these cities, an attempt has been made to download data. Two of them: Athens and Copenhagen, were discarded due to problems with geocoding the names to bounding polygons. For the rest of the cities, it was possible to download the data. However, as was mentioned earlier, some of the data in OpenStreetMap is incomplete and there is no easy automatic way to tell in which regions of the world this is the case. Therefore, for each of the remaining 54 cities, the data was manually browsed to assess its quality. The easiest way to do it, without knowing the cities, was to compare tagged buildings with satellite images. One can almost immediately see that there exist large areas without information on existing buildings. This kind of manual analysis only gave a rough estimate of the data quality, but it was enough to select 37 cities for the dataset. Most of the selected cities are marked on the maps in Figure \ref{fig:cities}.

\begin{table}[htpb]
\centering
\caption{OpenStreetMap keys used for analysis.}
\label{tab:osm-keys}
\begin{tabular}{|l|c|}
\hline
\textbf{Key}      & \textbf{Used} \\ \hline
aerialway         &               \\ \hline
aeroway           & \checkmark             \\ \hline
amenity           & \checkmark             \\ \hline
barrier           &               \\ \hline
boundary          &               \\ \hline
building          & \checkmark             \\ \hline
craft             &               \\ \hline
emergency         &               \\ \hline
geological        &               \\ \hline
healthcare        & \checkmark             \\ \hline
highway           &               \\ \hline
historic          & \checkmark             \\ \hline
landuse           & \checkmark              \\ \hline
leisure           & \checkmark             \\ \hline
man\_made         &               \\ \hline
military          & \checkmark             \\ \hline
natural           & \checkmark             \\ \hline
office            & \checkmark             \\ \hline
place             &               \\ \hline
power             &               \\ \hline
public\_transport &               \\ \hline
railway           &               \\ \hline
route             &               \\ \hline
shop              & \checkmark             \\ \hline
sport             & \checkmark             \\ \hline
telecom           &               \\ \hline
tourism           & \checkmark             \\ \hline
water             & \checkmark             \\ \hline
waterway          & \checkmark             \\ \hline
\end{tabular}
\end{table}

\subsection{Feature selection}
To obtain meaningful embedding vectors, the meaning must be present in the data used to create them. In order to ensure this, an initial feature selection was performed. The first filtering was done at the \textit{key} level of OpenStreetMap data. Out of 29 available keys, 15 were chosen for further analysis. The selections are shown in Table \ref{tab:osm-keys}. Most of the keys that were not included represent features that were thought to be irrelevant to an analysis of cities. For example, the \textit{aerialway} key is used to tag forms of transportation almost non-existent in cities, the \textit{barrier} key is used to show travelling barriers and obstacles and the \textit{boundary} key shows administrative borders irrelevant at the city level. 

We provide the following justification for the keys that seem controversial to reject: \textit{craft} was not used as it shows very specialized places that produce customized goods, and they were deemed irrelevant to the work, \textit{emergency} was discarded because it was mostly used by OSM users to tag fire hydrants only, and these are subject to poor tag coverage,\textit{highway}, \textit{public\_transport} and \textit{railway} were discarded, because analysis of these transportation systems was thought to be a complex task in itself and was not undertaken in this work.

The second part of filtering was performed at the \textit{value} level of the OSM data. Firstly, these values were explicitly discarded due to problems with tags coverage:
\begin{itemize}
    \item \textit{amenity}: \textit{waste\_basket},
    \item \textit{landuse}: \textit{grass},
    \item \textit{historic}: \textit{tomb},
    \item \textit{natural}: \textit{tree}, \textit{tree\_row}, and \textit{valley}.
\end{itemize}
\textit{Waste\_basket}, \textit{grass}, \textit{tomb}, \textit{tree} and \textit{tree\_row} were used inconsistently throughout the data set. For example, one can come across a problem where \textit{tree} and \textit{tree\_row} tags was performed very thoroughly, and where trees are present but there is no information about it. The \textit{tomb} tag was not used in the vast majority of the dataset. However, there was one cemetery in Warsaw where over 1000 individual graves were tagged. This introduced a lot of noise into the data, so the tag was discarded too. The \textit{valley} similarly did not appear in the data often, and in some cities also introduced a lot of noise so it was removed too. In the final step of the feature selection process, all tags that did not exist in any of the cities were removed. After that 725 distinct \textit{key}=\textit{value} pairs remained.

%\begin{figure}[H]
%    \centering
%    \includegraphics[width=0.48\textwidth]{images/powstancow.png}
%    \caption{Inconsistency in the use of the \textit{tree} tag on the example of Powstańców Śląskich Square in Wrocław.}
%    \label{fig:tree-powstancow}
%\end{figure}

\subsection{Dividing cities into regions}
\begin{figure}[h]
    \centering
    \includegraphics[width=0.48\textwidth]{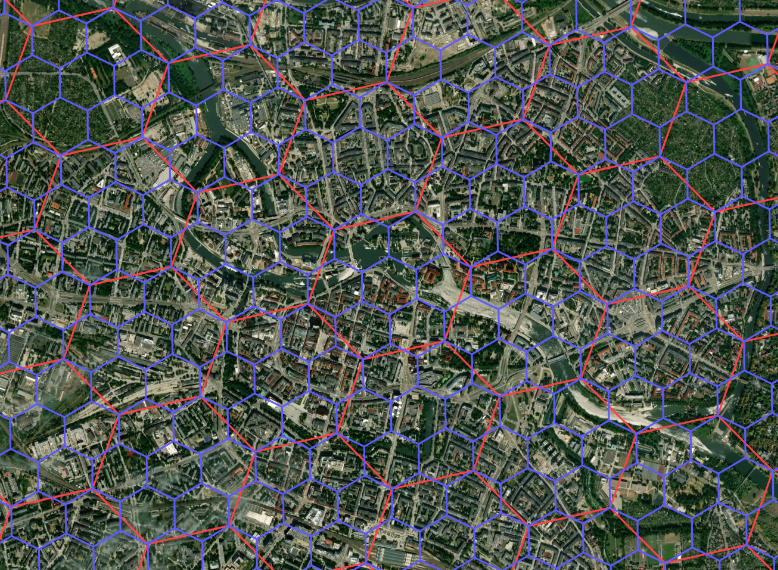}
    \caption{H3 cells of resolutions 8 (red) and 9 (blue) overlaid on top of a satellite image of Wrocław, Poland.}
    \label{fig:h3-8-9}
\end{figure}

In order to analyze cities at the regional level, a method of dividing them into regions had to be proposed. In this thesis, the H3 spatial index\cite{uber-h3} was used for this task. This made it possible to do this automatically, without good knowledge of the cities analyzed, maintaining the advantages mentioned in Section \ref{ch:spatial-indexes}. H3 spatial index divides the Earth's surface into a grid of hexagons at different resolutions. This allows analyses to be carried out at different levels of detail. Table \ref{tab:h3-resolutions} shows the cells' characteristics for a subset of available resolutions. The initial candidates were resolutions 8 and 9, as they have side lengths of around a few hundred meters and cover areas that correspond to reasonably small urban regions. Hexagons of these resolutions are presented in Figure \ref{fig:h3-8-9}. Finally, the resolution of 9 was chosen, as its hexagons cover the area of a small section of a neighborhood: several blocks and courtyards. This was the expected level of detail.

\begin{table}[h]
\centering
\caption{Cell characteristics for a subset of available H3 resolutions.\cite{h3-resolutions}}
\label{tab:h3-resolutions}
\begin{tabular}{@{}lrrr@{}}
\toprule
\multicolumn{1}{c}{\textbf{Resolution}} & \multicolumn{1}{c}{\textbf{Average Area [$\text{m}^2$]}} & \multicolumn{1}{c}{\textbf{Average Edge Length [m]}} \\ \midrule
7                                          & 5161293.2                                               & 1220.6                                             \\
8                                          & 737327.6                                               & 461.4                                               \\
9                                          & 105332.5                                               & 174.4                                               \\
10                                         & 15047.5                                               & 65.9                                                 \\
11                                         & 2149.6                                               & 24.9                                                  \\ \bottomrule
\end{tabular}
\end{table}

\subsection{Preprocessing: from geometries to feature vectors}\label{ch:tag-counting}
The final data transformation that needed to be performed was to transform the geometry data with tags attached into feature vectors for each region. A simple bag-of-words model was utilized here. First of all, each distinct \textit{key}=\textit{value} pair selected earlier was transformed into a feature named \textit{key\_value} forming $725$ such features. For each region delimited by a hexagon all elements from OpenStreetMap data, that spatially intersected with the region, were selected. After that, all tags of all the selected objects were counted forming a feature vector for that region. If an element intersected with two regions, it was counted in both of them. 

The last step was simple filtering. All regions with no elements in them (all features equal to $0$) were discarded. After performing this step, a feature matrix of size $183961 \times 725$ was obtained. Each row of the matrix represented a feature vector for a certain region, and each column showed the number of elements with a certain tag in all regions. 
The number of regions obtained for each city is presented in Figure \ref{fig:city-number-regions}. To have some idea of the difference in the scale of tagging in different cities, an additional analysis was carried out. This was done at the level of keys, as there are too many values to visualise.

First, regions in which each feature was non-zero, i.e. there was at least one object with the given feature, were counted. Then, for each city, the percentage of regions containing a given feature was counted. Finally, these percentages were normalized to allow them to be presented on a heatmap. In the columns, you can see the differences that may result from differences in tagging or the characteristics of the city. This is the case, for example, for Paris. In many columns, but especially in healthcare, it significantly stands out from the rest, which may suggest differences in tagging.
In turn, rows that appear empty suggest a potential problem with tag coverage. This is the case for Helsinki and Reykjavik. However, after manual verification, it turned out that in the case of both of these cities there is a lot of undeveloped land within their borders, such as mountains, forests and parts of the sea. After checking these two cases, the obvious problems with the data can no longer be easily seen from observing the heatmap.

%\begin{figure}[htpb]
%    \centering
%    \includegraphics[width=0.48\textwidth]{images/nonzero_tags_heatmap.pdf}
%    \caption{Normalized percentages of regions with non-zero numbers of each OpenStreetMap key in them. High values mean that a higher percentage of regions in a city had at least one element from a given category, compared to the rest of the cities.}
%    \label{fig:nonzero-heatmap}
%\end{figure}

\section{Proposed solution}
We propose a new OpenStreetMap embedding method -- Hex2Vec. Firstly, we discuss the way regions are divided will be recalled. Then, the theoretical description of the proposed method and the learning procedure will be presented. Later, the connection with solutions based on triplet networks will be commented on. Finally, a solution for region similarity detection will be presented.

\subsection{Division into regions}
As stated before, Uber's H3 spatial index was used to create regions in this thesis. The index is described in more detail in Chapter \ref{ch:spatial-indexes} while the division and data pre-processing is described in Chapter \ref{ch:data}.

Theoretically, the proposed method is generic and should work with any division of analyzed space. However, H3 has the following features which, according to the author of this thesis, can have a positive impact on the quality of the obtained embeddings:
\begin{itemize}
    \item similar size of the obtained regions,
    \item equal distance between the centers of the regions,
    \item no distinction between types of neighbors (e.g. through vertex and edge as in the case of square regions),
    \item efficient approximation of a circular neighborhood.
\end{itemize}

\subsection{Proposed method}
The method proposed in this thesis, called Hex2Vec, is based on a model from the well-established domain of natural language processing. Namely, it is based on the Skip-gram model with negative sampling. It was proposed by Mikolov et al. in \cite{word2vec-n} as a solution to learning word embeddings. Its description provided here is based on the scheme presented in \cite{fasttext}, because it is clear and its use simplifies understanding.
\subsection{Skip-gram with negative sampling}
Given the training corpus, represented as a sequence of words $w_1, w_2, \ldots, w_T$, the general objective of the Skip-gram model is to maximize the following log-likelihood:
\begin{equation}
    \sum_{t=1}^T\sum_{c\in C_t}\log p(w_c | w_t),
\end{equation}
where $C_t$ is the set of indices forming the context of the word $w_t$. Originally\cite{word2vec} the probability $p(w_c | w_t)$ was parametrized using hierarchical softmax, however, it was replaced with negative sampling in later works\cite{word2vec-n}. To do so, the problem was framed as a set of binary classification tasks. In order to describe it, let us assume that we have a function $s: (w_c, w_t) \mapsto \mathbb{R}$ that maps pairs of words to scores in the set of real numbers. Then we obtain the negative log-likelihood for one pair of words $(w_c, w_t)$:
\begin{equation}
    \log\left(1 + e^{-s\left(w_c, w_t\right)}\right) + \sum_{n \in N_{t,c}} \log\left(1 + e^{s\left(w_n, w_t\right)}\right),
\end{equation}
where $N_{t,c}$ denotes a set of negative samples, formed by randomly sampling words from the dictionary. The general idea is therefore to create a model able to distinguish between the true (word, context) pairs and ones drawn from a random noise distribution. Hence the use of binary cross entropy as the minimized loss function.

\subsection{Hex2Vec}
To apply the Skip-gram model to geospatial data, an assumption is made that it is possible to predict well whether given two regions are neighbors or not. This assumption is derived from \textit{the first law of geography} \cite{first-law}, which is commonly mentioned in the literature. 

\begin{figure}[htpb]
    \centering
    \includegraphics[width=0.48\textwidth]{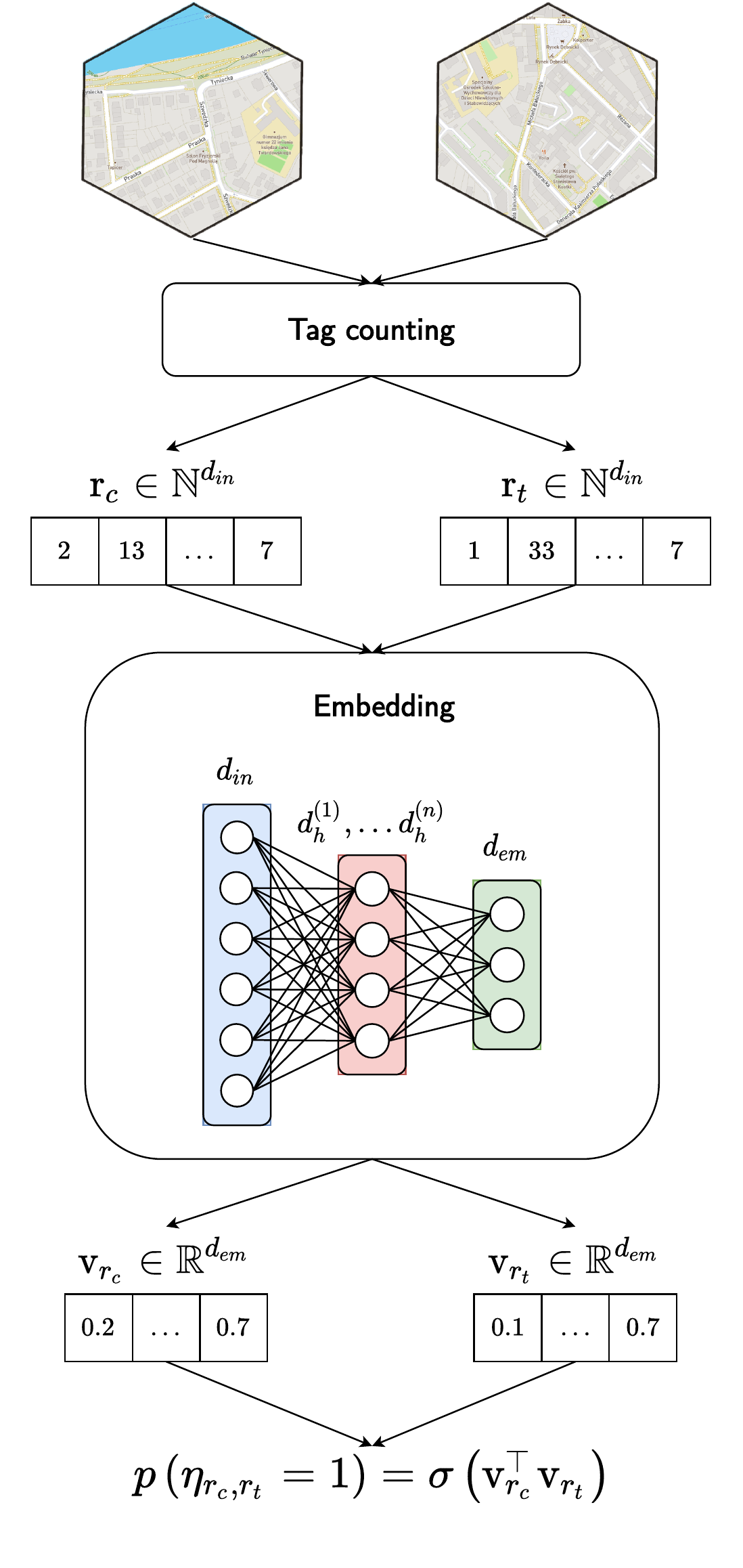}
    \caption[Hex2Vec workflow diagram.]{Hex2Vec workflow diagram. $d_{in}$ is the size of the input vector after tag counting, $d_{em}$ is the size of the embedding vector, $d_{h}^{(i)}$ is the size of the network's $i$-th hidden layer. $\mathbf{r}_c$ and $\mathbf{r}_t$  denote the vectors obtained from tag counting while $\mathbf{v}_{r_c}$ and $\mathbf{v}_{r_c}$ denote their low-dimensional representation.}
    \label{fig:model}
\end{figure}
After that, it is very straightforward to adapt the Skip-gram model to city regions. Let's start with defining the scoring function:
\begin{equation}
    s(r_c, r_t) = \mathbf{v}_{r_c}^\top \mathbf{v}_{r_t},
\end{equation}
and the probability of neighborhood:
\begin{equation}\label{eq:probability}
    p\left(\eta_{r_c, r_t} = 1\right) = \sigma\left(s\left(r_c, r_t\right)\right),
\end{equation}
where $\eta_{r_c, r_t}$ is a binomial random variable such that:
\begin{equation}
    \eta_{r_c, r_t} = \begin{cases}
    1& \text{if } r_c \text{ and } r_t \text{ are neighbors}\\
    0              & \text{otherwise}.
\end{cases}
\end{equation}
The $r_t$ and $r_c$ symbols are used to denote the input region and its context, while $\mathbf{v}_{r_t}$ and $\mathbf{v}_{r_c}$ denote their low-dimensional vector representations in the embedding space. In the original Skip-gram model, these were parameterized using word vectors. However, it would be impossible in this case, as there is no closed dictionary of regions. Therefore, tag counting was first performed for each region as described in section \ref{ch:tag-counting}. Then a fully connected neural network was used to project the obtained high-dimensional region vector $\mathbf{r}_i$ to its representation in the embedding space $\mathbf{v}_{r_i}$. The whole process is shown in Figure \ref{fig:model}.

One further change was made to the original objective. Namely, only one negative sample $r_n$ was drawn for each pair $(r_c, r_t)$. Taking this into account, and denoting the logistic loss function as $l: x \mapsto 
\log\left(1 + e^{-x}\right)$ the final Hex2Vec objective is obtained:
\begin{equation}
    \sum_{t=1}^T \left[ \sum_{c \in C_t} l\left(s \left(r_c, r_t\right)\right) + \sum_{n \in N_t} l\left(-s \left(r_n, r_t\right)\right) \right],
\end{equation}
where:
\begin{itemize}
    \item $T$ is the number of all regions in the dataset,
    \item $C_t$ is the set of indices of context regions of region $r_t$,
    \item $N_t$ is the set of indices of negative samples of region $r_t$.
\end{itemize}
The last thing that needs explaining is how the context and the negative samples are obtained. This will be presented next.

\subsection{Context and negative sampling}
\begin{figure}[htpb]
    \centering
    \includegraphics[width=0.48\textwidth]{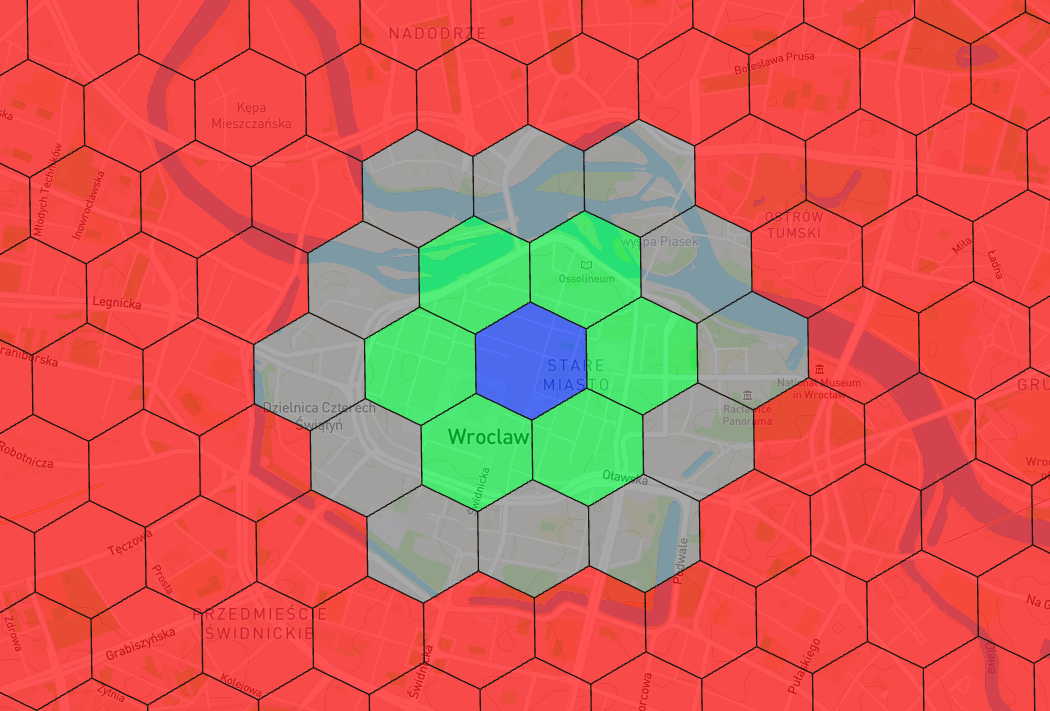}
    \caption{Sampling regions represented visually. The context of the blue region is shown in green. The negative samples are chosen randomly from the regions in red. The regions from the second ring, shown in grey, were not used for this region.}
    \label{fig:sampling}
\end{figure}

To complete the embedding method definition the notion of context and the negative sampling process has to be defined. Both of these are simple and follow what can be seen in the literature. For the context of a region, its first neighbors are selected. This is where the advantages of H3 are noticeable as the context efficiently estimates a circular neighborhood, and each neighbor is the same distance away from the centre. The negative samples on the other hand are drawn at random from outside the 2 rings around a given region. This is presented in Figure \ref{fig:sampling}.

\section{Similarity detection}
After training the embedding model, and obtaining the regions' vector representations, similarity detection of the regions will be performed. This will be done using a bottom-up hierarchical approach, namely agglomerative clustering. This way a multi-level division of regions in the city will be obtained, which will then be described level by level, thus proposing a typology of regions. 
Later on, pairwise similarity will also be used to further explore the embedding space.
\subsection{Agglomerative clustering}
The agglomerative clustering method works in a bottom-up fashion. It starts with each data point in its own cluster and then merges pairs of clusters according to a selected distance measure and linkage criterion.
The distance measure is used to determine distances between pairs of data points.
Two measures used in this thesis will be the Euclidean distance:
\begin{equation}
    d_e\left(\mathbf{x}, \mathbf{y}\right) = || \mathbf{x} - \mathbf{y} ||
\end{equation}
and the cosine distance:
\begin{equation}
    d_c\left(\mathbf{x}, \mathbf{y}\right) = 1 - \frac{\mathbf{x} \cdot \mathbf{y}}{||\mathbf{x}|| ||\mathbf{y}||},
\end{equation}
where $||\mathbf{x}||$ denotes the Euclidean norm of the vector $\mathbf{x}$.

After the first merging of two clusters, a simple distance measure can no longer be used because one cluster contains two observations. The linkage criterion is then applied. It defines the distance between clusters using the distance between pairs of observations from the clusters. Multiple such criteria have been proposed, but in this thesis, Ward's method\cite{ward} will be used. At each step, it merges two clusters in a way that minimizes the increase in total intracluster variance after merging. This criterion tends to generate strongly connected clusters where objects are very similar to each other. This makes it a good choice for pattern analysis with the intent to define a typology of regions. This criterion forces the use of the Euclidean distance, so this measure will be used in agglomerative clustering.
The cosine distance, on the other hand, will be used in the analysis of the pairwise similarity between regions. Due to normalisation, it should denote two vectors as similar if they share certain features, even if those features have different magnitudes.

\subsection{Region typology}
This section presents the results of similarity detection carried out for regions in 6 Polish cities. It was performed using hierarchical clustering and embedding vectors from the proposed Hex2Vec model. The results were discussed in a top-down fashion, split by split. 
The obtained representations allowed to separate regions quite well based on their functions. One of the clusters was not fully divided, but from a certain level, the clustering became significantly noisier. It was therefore decided to stop the analysis at the level of 10 clusters. The obtained clusters can be described as follows:
\begin{enumerate}
    \addtocounter{enumi}{-1}
    \item not fully divided cluster that contains mainly meadows, but also allotments and some regions with single-family housing,
    \item regions with very good availability of various leisure facilities and amenities. The cluster includes mainly commercial city-centre regions, but also some regions with multi-family housing and service regions on single-family housing estates,
    \item residential areas with mainly single-family housing located on agricultural land. Good examples are Warsaw districts such as Białołęka, Wawer and part of Ursynów,
    \item agricultural land with sparse development,
    \item areas of forests,
    \item nature reserves,
    \item airports,
    \item single-family residential areas not located on agricultural land,
    \item rivers,
    \item industrial areas.
\end{enumerate}

Looking at the results, it can be noticed that the embedding model captures well large regions that span larger neighborhoods such as city centres, suburban residential regions or nature reserves. Some problems arise in differentiating urban regions with single-family housing. In the analysis, some such regions were not separated from cluster 0 and some were included in cluster 7. However, the most confusion arises in the latter grouping when small clusters are formed. In part, this may be due to differences in the quality of building tagging. On the other hand, given how the model is learned, the higher quality of differentiation of larger regions is not surprising. Further work could therefore be to improve the model to also include the interior of the region in the representation. The obtained are not perfect but are nevertheless satisfactory and allow to draw preliminary conclusions about the structure of cities even without prior knowledge.

\section{Operations in the embedding space}
Performing vector arithmetic and interpolation in embedding space is intended to show that these operations work intuitively as we would expect them to due to the semantics of the manipulated objects, and are often used as methods of evaluating representations. Mikolov et al. for example showed that in the case of word embeddings it was possible to perform operations such as $vector("King") - vector("Man") + vector("Woman")$ with the result being closest to $vector("Queen")$. Both vector arithmetic and interpolation was also used in works tackling geospatial data embedding such as Loc2Vec\cite{loc2Vec} and Tile2Vec\cite{tile2vec} achieving satisfactory results. A similar experiment was performed as a part of this thesis, and the results are presented in this section.

\subsection{Vector addition}
\begin{figure}[H]
    \centering
    \includegraphics[width=0.48\textwidth]{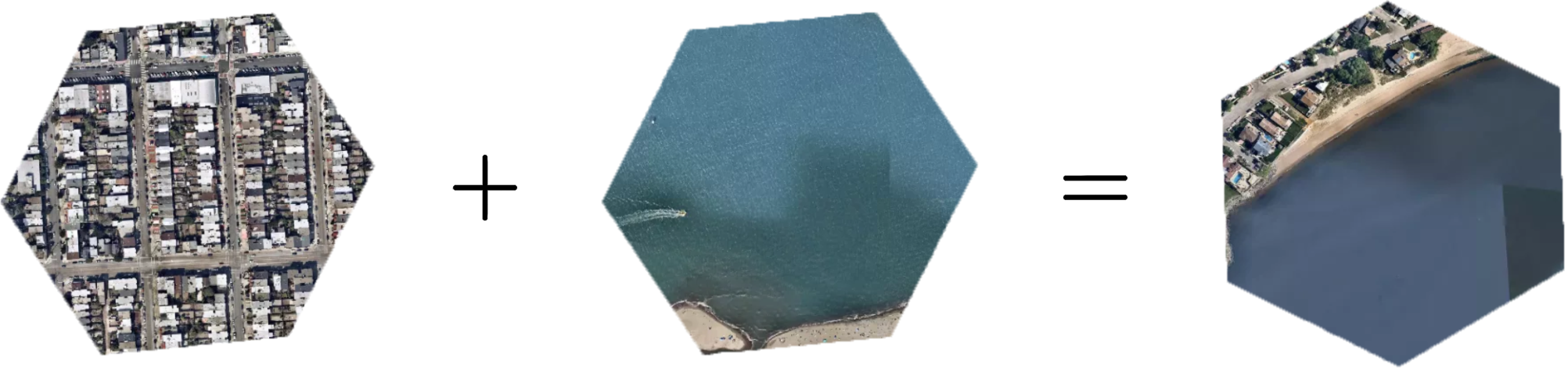}
    \caption[Vector addition in the embedding space]{Vector addition in the embedding space. Adding a region with a coast to a densely built-up region results in a region with buildings along the coast.}
    \label{fig:addition-coast}
\end{figure}

To perform vector addition, two regions were chosen in the city of San Francisco. One region was a densely built-up commercial area, and the other one was a coastal region without buildings. These two regions were embedded and their representations were added together. After that, another region was found in the dataset that was closest to the result according to cosine similarity. This turned out to be a region in New York City with buildings along the coast. The process is presented in Figure \ref{fig:addition-coast}.

% \begin{figure}[htpb]
%      \centering
%      \begin{subfigure}[b]{\textwidth}
%          \centering
%          \includegraphics[width=0.48\textwidth]{images/city_plot.png}
%          \caption{Features of the city region}
%          \label{fig:nonzero-city-region}
%      \end{subfigure}
%      \hfill
%      \begin{subfigure}[b]{\textwidth}
%          \centering
%          \includegraphics[width=0.48\textwidth]{images/beach_plot.png}
%          \caption{Features of the coastal region}
%          \label{fig:nonzero-coast}
%      \end{subfigure}
%      \hfill
%      \begin{subfigure}[b]{\textwidth}
%          \centering
%          \includegraphics[width=0.48\textwidth]{images/between_plot.png}
%          \caption{Features of the closest region after addition}
%          \label{fig:nonzero-coast-result}
%      \end{subfigure}
%         \caption{Non-zero tag counts for beach region, city region and the closest one after addition.}
%         \label{fig:nonzero-tags-addition}
% \end{figure}

\subsection{Vector subtraction}
\begin{figure}[H]
    \centering
    \includegraphics[width=0.48\textwidth]{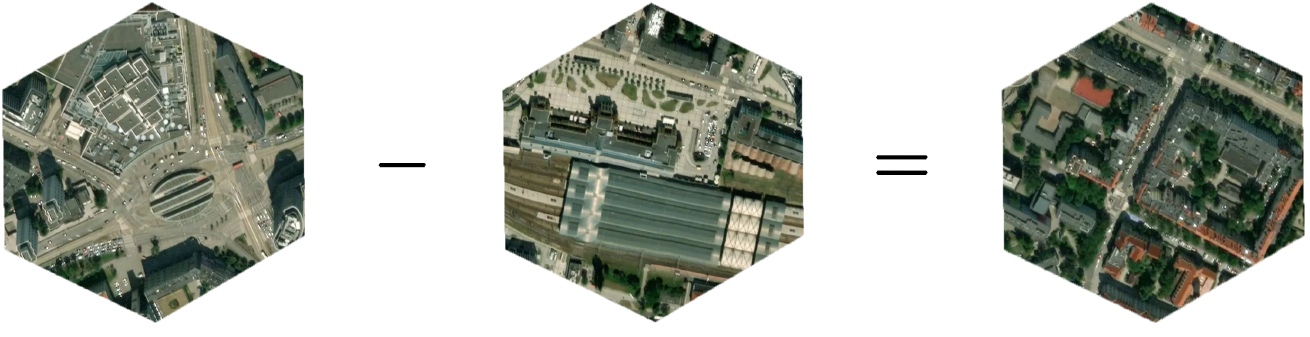}
    \caption[Vector addition in the embedding space]{Vector subtraction in the embedding space. Subtracting a region surrounding the train station from a region with a shopping centre and a university, results in a region with university and school.}
    \label{fig:subtraction-pwr}
\end{figure}

To test the operation of subtraction, a less trivial example was chosen. Two regions of Wroclaw were selected: Grunwaldzki Square and the main railway station. Their representations were subtracted and the nearest region was found according to cosine similarity. Initially, the expected result was some region with a shopping centre.
The result turned out to be surprising although sensible. It turned out that the region of Grunwaldzki Square contains part of the buildings of the Wroclaw University of Technology. Thus, the most influential difference turned out to be the presence of the university. The final result, therefore, turned out to be a different region of the Wrocław University of Technology campus - as in the subtraction operation in Figure \ref{fig:subtraction-pwr}.

%\begin{figure}[htpb]
%     \centering
%     \begin{subfigure}[b]{\textwidth}
%         \centering
%         \includegraphics[width=0.48\textwidth]{images/grunwald_dworzec_diff.pdf}
%         \caption{Difference between the shopping centre/university region and the train station region}
%         \label{fig:difference-grunwald-dworzec}
%     \end{subfigure}
%     \begin{subfigure}[b]{\textwidth}
%         \centering
%         \includegraphics[width=0.48\textwidth]{images/pwr_result.pdf}
%         \caption{Resulting university region's tag counts}
%         \label{fig:pwr-result}
%     \end{subfigure}
%     \caption{Difference in tag counts between the shopping centre/university region and the train station region (\subref{fig:difference-grunwald-dworzec}) and the resulting university region's tag counts (\subref{fig:pwr-result})}
%    \label{fig:pwr-subtraction-features}
%\end{figure}

% \paragraph{Manually crafted feature vector}

\subsection{Interpolation in the latent space}
\begin{figure}[htb]
    \centering
    \includegraphics[width=.48\textwidth]{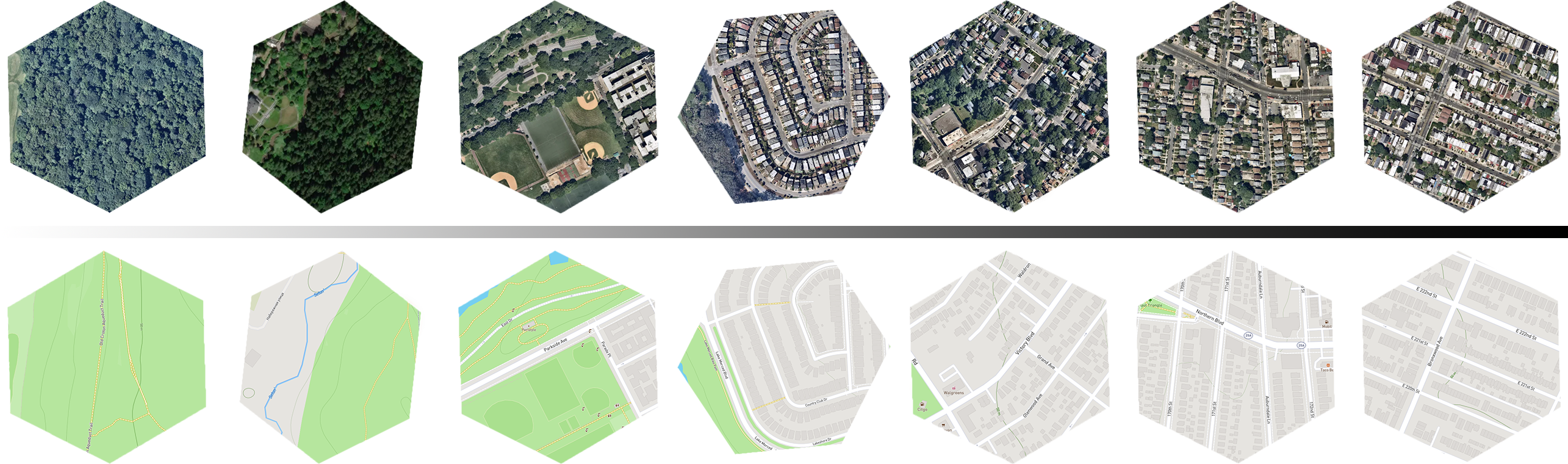}
    \caption[Interpolation in the embedding space between two regions in New York City]{Interpolation in the embedding space between two regions in New York City. The initial region is in a park and the final region is in a dense residential area.}
    \label{fig:interpolation}
\end{figure}

To perform interpolation two regions from New York City were chosen. The first one contained only a park, while the other one was a densely built-up residential region. After embedding, a linear interpolation was performed yielding five intermediate vectors between the two. The process is shown in Figure \ref{fig:interpolation} both using satellite imagery and OpenStreetMap rasters. It can be seen that the regions gradually move towards more buildings until they no longer contain a park. This process is also in line with human intuition, which may indicate that good representations were learned for these regions.

\section{Conclusions and Future Work}
We proposed a method for unsupervised embedding of urban micro-regions based on functional information available in OpenStreetMap which allows region similarity detection based on the obtained embeddings. We thus fill a research gap in using OpenStreetMap data for micro-region embedding highlighted in the background section. Afterwards, a dataset consisting of regions from 37 cities was created. As part of this step, manual selection of cities was carried out, feature selection was performed and a method was proposed to group the data into regions using the H3 spatial index. Next, a new method for embedding micro-regions using OpenStreetMap data, called Hex2Vec, was proposed. To the best of our knowledge, it is the first method of micro-region embedding, using numerical data from OpenStreetMap. The Hex2Vec model was then trained and used to obtain the embeddings for regions from the dataset. The embeddings were used to obtain high-level typology of cities' micro-regions using hierarchical clustering. Lastly, the vector arithmetic and interpolation in the latent space were presented. It was shown that these operations have the expected semantics. 

We hope that hex2vec inspires further research in the area of geospatial embedding of OSM data. A variety of research directions are available, among others aspects that were not taken up in this work such as: using obtained embeddings in downstream tasks or taking more OSM tags into consideration. Other improvements are natural extensions coming from other fields of representation learning such as using subword information and attention-like mechanisms to improve both the detail of what is represented and to move further in context from the local scale. Additionally softer mining strategies and tessellation methods could be investigated. 

\bibliographystyle{ACM-Reference-Format}
\bibliography{sample-base}

\newpage
\appendix

\section{Data set statistics}

\begin{figure}[H]
    \centering
    \includegraphics[width=0.48\textwidth]{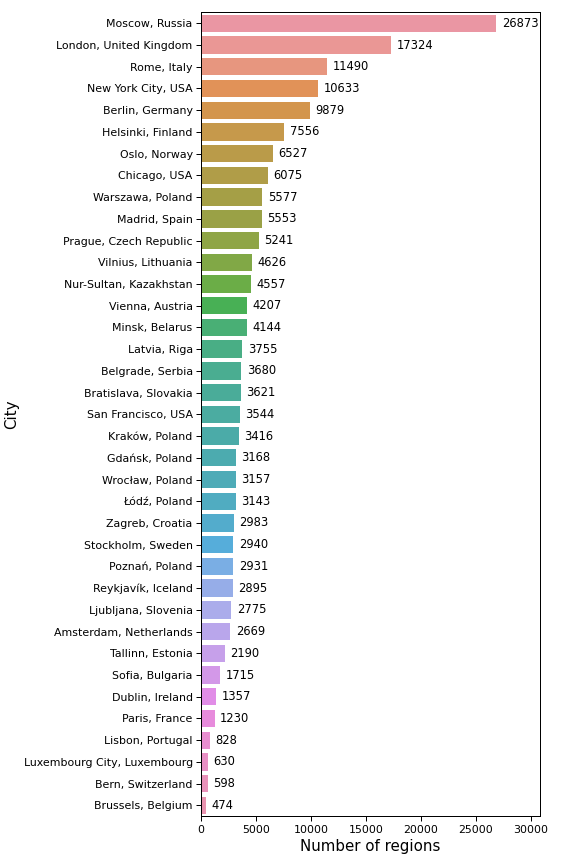}
    \caption{Number of regions in each city.}
    \label{fig:city-number-regions}
\end{figure}

\end{document}